# Diverse Linguistic Features for Assessing Reading Difficulty of Educational Filipino Texts

**Joseph Marvin IMPERIAL[a,b] & Ethel ONG[b]**
[a]*National University, Manila, Philippines*
[b]*De La Salle University, Manila, Philippines*
jrimperial@national-u.edu.ph[a], ethel.ong@dlsu.edu.ph[b]

**Abstract:** In order to ensure quality and effective learning, fluency, and comprehension, the proper identification of the difficulty levels of reading materials should be observed. In this paper, we describe the development of automatic machine learning-based readability assessment models for educational Filipino texts using the most diverse set of linguistic features for the language. Results show that using a Random Forest model obtained a high performance of 62.7% in terms of accuracy, and 66.1% when using the optimal combination of feature sets consisting of traditional and syllable pattern-based predictors.

## 1. Introduction

Reading is a vital life skill that children learn early during their primary years. It is key towards gaining good academic performance, as the ability to read not only affects language and literacy development, but also the ability to understand and solve math problems and comprehend science facts. Learning to read entails multiple processes that include word recognition (Imperial and Ong, 2021), comprehension, and fluency in order to make meaning from the printed text (Klauda and Guthrie, 2008). Thus, the proper identification of readability levels of children's literature is very important. Previous studies have reported that children who read educational materials such as story and picture books often experience frustration and boredom when the reading level is not within their range of ability (Gutierrez, 2015; Guevarra, 2011). However, readability assessment comes with a challenge—manual extraction of linguistic variables such as determining occurrences of polysyllabic words, calculating average character length, tagging grammatical labels of words and plugging them to handcrafted formulas can be heavily time-consuming and tedious. In addition, linguistic variables that can potentially affect the difficulty levels of texts vary from language to language. In this study, we investigate the performance of other possible factors affecting readability from a language's characteristics such as lexical, language model, syllabication rules, and morphology for readability assessment of educational text in Filipino.

## 2. Related Work

While high-resource languages such as English (Vajjala and Lučić, 2018; Collins-Thompson and Callan, 2014; Feng et al., 2010; Collins-Thompson and Callan, 2005) and German (Hancke and Meurers, 2013; Hancke et al., 2012) boast a rich history of research efforts in readability assessment, the same cannot be said for low-resource languages where machine-readable data is limited and NLP-based tools for extracting complex linguistic predictors are yet to be developed. To compare, research on high-resource languages have explored more potential linguistic predictors, even extending to psycholinguistics (Vajjala and Meurers, 2014), cognitive-based features (Crossley et al., 2008) and gaze movement (Gonzalez-Garduno and Søgaard., 2017), than low-resource languages such as Filipino, Bangla, Telugu, and other South Asian languages.

In Filipino, experts conducting research in the field such as Macahilig (2015) and Gutierrez (2015) have outlined possible linguistic factors that should be considered specifically for the language. These factors include the lexical categories of words, syllable patterns, syntactic features, and text structure. To date, few research works have explored using these prescribed features. Guevarra (2011) initiated the use of a machine learning model via logistic regression with seven traditional features such as number of unique words, average number of syllables, and total number of sentences. Later works by Imperial et al. (2019), and Imperial and Ong (2020, 2021) explored the inclusion of lexical features such as lexical densities, foreign and compound word densities, as well as language model features such as word and character trigrams and noted significant improvement to performance when using the latter. We further extend these works by supplying linguistic predictors that have not been explored yet such as features on syllable patterns based on the orthography of the Filipino language and morphological features which have never been used before in Filipino text readability research, thus, completing the recipe for text-based features for readability assessment of Filipino texts.

## 3. Educational Filipino Text Corpus

The Filipino text corpus used for training and validating our ML models is derived from two sources: *Adarna House Corpus* and *DepEd Commons*. Our study focuses only on resources for early grade learners as this is the group where reading materials are abundant online.

**Adarna House Corpus.** We obtained permission from Adarna House Inc., the largest children's literature publisher in the Philippines accredited by the Department of Education, to use their machine-readable copy of leveled reading materials. A total of 174 reading materials were obtained, spanning from grades 1 to 3 in the form of fictional story books written in Filipino. This resource has been pre-annotated by Adarna's in-house language experts and writer-researchers in terms of readability level.

**DepEd Commons.** DepEd Commons is an online platform launched by the Department of Education (DepEd) which contains open-source books and reading materials for all basic education subjects in various grade levels such as *English, Filipino, Science, Araling Panlipunan* (Social Studies), *Arts*, and *Physical Education*. These are available for free download by public and private school teachers and learners. A total of 91 reading passages in the form of short stories also spanning from grades 1 to 3 were extracted from Filipino activity books obtained from the repository and added to the current data count with the Adarna House.

## 4. Diverse Linguistic Features

The concept of automating the readability assessment task draws from the notion of extracting a wide range of linguistic features often recommended by experts that can potentially affect the readability of texts (Macahilig, 2015; Gutierrez, 2015). As such, we performed extraction of linguistic feature sets from the corpus as numerical vector representations required for model training. A total of **54 linguistic predictors** from 5 different feature sets covering various facets of texts such as **surface-based or traditional (TRAD), lexical (LEX), language structure (LM), syllable pattern (SYLL)**, and **morphological predictors (MORPH)** were used for this study. To note, this is the most extensive number of linguistic factors considered for Filipino readability assessment to date, much more than features extracted from the works of Macahalig (2015), Imperial et al. (2019), and Imperial and Ong (2020) with 15, 15, and 25 predictors respectively. Table 1 describes each predictor from the extracted feature sets.

Table 1. *Breakdown of various linguistic predictors per feature set.*

| Feature Set | Count | Predictors |
| --- | --- | --- |
| **TRAD** - Traditional or surface-based features based on counts and frequencies. | 7 | Total word, sentence, phrase, polysyllabic word counts. Average word length, sentence length, syllable per word. |
| **LEX** - Lexical or context carrying features via part-of-speech categories. | 9 | Type-token variations (regular, logarithmic, corrected, root). Noun and verb token ratio. Lexical density. Foreign word and compound word density. |
| **LM** - Language model features based on perplexity. | 9 | Language models trained on three levels (L1, L2, and L3) of the external DepEd Commons corpus using n-gram values of {1, 2, 3}. |
| **SYLL** - Syllable pattern densities based on the prescribed national orthography. | 10 | Consonant cluster density. Densities of the prescribed Philippine orthography on syllable patterns: {*v, cv, vc, cvc, vcc, cvcc, ccvc, ccvcc, ccvccc*} where *c* and *v* are consonant and vowel notations. |
| **MORPH** - Morphological features based on verb inflection. | 19 | Densities of various foci of verbs based on tense: {actor, object, benefactive, locative, instrumental, referential}. Densities of various foci of verbs based on aspect: {infinitive, perfective, imperfective, contemplative, participle, recent-past, auxiliary}. |

## 5. Experiment Setup

To develop models for automatic readability assessment, we select commonly-used machine learning algorithms for document classification, namely Random Forest (RF), and Support Vector Machines (SVM), and use the extracted feature sets from the reading materials and their corresponding labels. During actual training, *k*-fold cross validation was performed with $k = 5$ for each algorithm. The performance of the models were evaluated using accuracy, precision, recall, and F1 score.

## 6. Performance Results of Machine Learning Models

We conducted a series of model training using the extracted linguistic feature sets to empirically understand and analyze each selected machine learning algorithm's performance on the automatic readability assessment task. Each feature set is selected and modeled singularly followed by combining feature sets together incrementally until all combinations are used. All models from these experiments have their hyperparameters optimized through exhaustive grid search.

Table 2. *Ablation experiments of model training using the extracted linguistic feature sets for **Support Vector Machines** with hyperparameters optimized. The top performing combination of features is highlighted in **boldface**.*

| Feature Set | Accuracy | Precision | Recall | F1 Score |
|---|---|---|---|---|
| TRAD | 0.475 | 0.460 | 0.475 | 0.465 |
| LEX | 0.424 | 0.421 | 0.424 | 0.415 |
| SYLL | 0.458 | 0.447 | 0.458 | 0.449 |
| LM | 0.390 | 0.369 | 0.390 | 0.373 |
| MORPH | 0.322 | 0.366 | 0.322 | 0.308 |
| TRAD + LEX | 0.475 | 0.449 | 0.475 | 0.406 |
| **TRAD + LM** | **0.542** | **0.532** | **0.542** | **0.536** |
| TRAD + SYLL | 0.508 | 0.484 | 0.508 | 0.490 |
| TRAD + MORPH | 0.492 | 0.476 | 0.492 | 0.465 |
| TRAD + LEX + SYLL | 0.475 | 0.452 | 0.475 | 0.460 |
| TRAD + LEX + LM | 0.475 | 0.475 | 0.475 | 0.475 |
| TRAD + LEX + MORPH | 0.475 | 0.456 | 0.475 | 0.456 |
| TRAD + LEX + SYLL + LM | 0.492 | 0.494 | 0.492 | 0.492 |
| TRAD + LEX + SYLL + MORPH | 0.492 | 0.499 | 0.492 | 0.495 |
| ALL | 0.492 | 0.481 | 0.492 | 0.485 |

From the training results of SVM using the singular feature sets in Table 2, TRAD emerged as the top single-feature predictor amongst all other feature sets. In addition, using MORPH feature sets obtained the lowest from the five. This result further strengthens the importance of TRAD features in readability assessment as done previously. The best performing model from the Support Vector Machine experiments uses a combination of **TRAD + LM** feature sets with an accuracy of 0.542, precision of 0.532, recall of 0.542, and F1 score of 0.536. With LM features present in all top-performing models, one inference that can be made from this is that the language model features are often the most-used support vectors for discriminating readability levels between classes with respect to how Support Vector Machine works.

Table 3. *Ablation experiments of model training using the extracted linguistic feature sets for **Random Forest** with hyperparameters optimized. The top performing combination of features is highlighted in **boldface**.*

| Feature Set | Accuracy | Precision | Recall | F1 Score |
|---|---|---|---|---|
| TRAD | 0.525 | 0.528 | 0.525 | 0.526 |
| LEX | 0.508 | 0.487 | 0.508 | 0.492 |
| SYLL | 0.525 | 0.530 | 0.525 | 0.515 |
| LM | 0.458 | 0.448 | 0.458 | 0.452 |
| MORPH | 0.475 | 0.494 | 0.475 | 0.475 |
| TRAD + LEX | 0.593 | 0.588 | 0.593 | 0.588 |
| TRAD + LM | 0.610 | 0.594 | 0.610 | 0.597 |
| **TRAD + SYLL** | **0.661** | **0.651** | **0.661** | **0.640** |
| TRAD + MORPH | 0.576 | 0.564 | 0.576 | 0.563 |
| TRAD + LEX + SYLL | 0.644 | 0.634 | 0.644 | 0.634 |
| TRAD + LEX + LM | 0.610 | 0.573 | 0.610 | 0.583 |
| TRAD + LEX + MORPH | 0.525 | 0.537 | 0.525 | 0.529 |
| TRAD + LEX + SYLL + LM | 0.627 | 0.602 | 0.627 | 0.605 |
| TRAD + LEX + SYLL + MORPH | 0.593 | 0.573 | 0.593 | 0.571 |
| ALL | 0.627 | 0.623 | 0.627 | 0.624 |

Using the singular feature sets for Random Forest, the TRAD emerged as one of the feature sets used by the top-performing models similar to Support Vector Machines. Thus, the efficacy of TRAD features is substantial for readability assessment of Filipino texts regardless of what machine learning algorithm is used. This result can also serve as a further basis that the use of traditional features from previous works (Villamin & de Guzman, 1979; Gutierrez, 2015; Macahilig, 2015) is practical. The other feature set that contributed towards obtaining the best performance is the SYLL feature set or predictors using syllable patterns (**TRAD + SYLL**). We infer that the syllable patterns can also substantially influence word difficulty in Filipino whereas more common patterns such as *cvc, vc,* and *cv* appear on highly readable texts and uncommon patterns such as *ccvcc* and *ccvccc* are more common on advanced texts. It is also of no surprise that the combination of these two feature sets would result in the top-performing model with an accuracy of 0.661, precision of 0.651, recall of 0.661, and F1 score of 0.640 among all models. For scale, the 0.661 accuracy is the current highest score for a readability assessment model in Filipino up to date from this work on studying the most number of linguistic features.

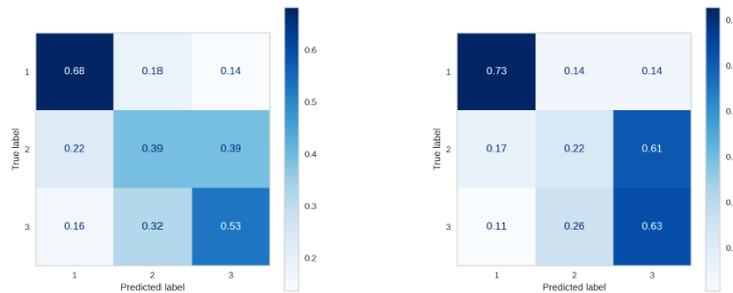

*Figure 1.* Confusion matrices of top models using *all* 54 extracted features (left) and *optimal combination* of features (TRAD + LM) for Support Vector Machines (right).

We also provide an analysis of correctly classified instances and fail cases via confusion matrices for the trained readability models. Figure 1 depicts the two confusion matrices for the model using all the feature sets (left) and the top-performing model using TRAD + LM (right) for comparison for Support Vector Machines. From the figures, both models often confuse the readability levels between Grade 2 and Grade 3. This observation is more prominent in the case of the TRAD + LM model where the shade is darker. Though it obtained the highest scores for the evaluation metrics, it does not perform well when classifying the Grades 2 and 3 text compared to the model using all feature sets. One inference from this is the decrease in the number of features (38 features not used) in TRAD + LM caused some confusion when differentiating Grade 2 from Grade 3. We note that there is a slight *trade-off* between the use of all features versus the optimal combination on the performance of models.

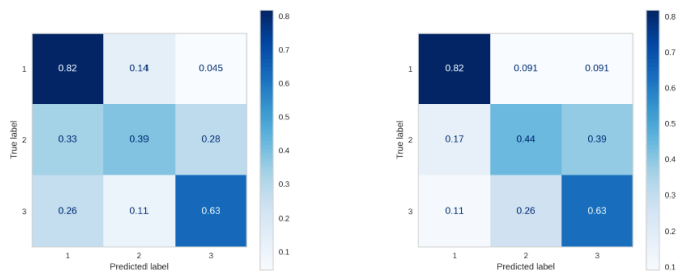

*Figure 2.* Confusion matrices of top models using *all* 54 extracted features (left) and *optimal combination* of features (TRAD + SYLL) for Random Forest (right).

Figure 2 describes the two confusion matrices for the model using all the feature sets (left) and the top-performing model using TRAD + SYLL (right) for Random Forest. From the figures, the majority of the predicted labels were matched to their corresponding true labels and both also performed better when classifying Grades 1 and Grades 3 from other grade levels. The confusion of discriminating between Grades 2 and 3 from the models developed using Support Vector Machines is not exhibited using Random Forest. This can be traced to the nature of training Random Forest using an ensemble of decision trees using random feature subsets by a number of tree estimators to avoid bias and reduce overfitting (Ho, 1995).

## 7. Conclusion

In order to ensure quality and effective learning and comprehension, the proper identification of the difficulty levels of reading materials should be observed. In this study, an automatic readability assessment model was trained from children's education materials using deep, hybrid linguistic feature sets spanning traditional, lexical, language model, syllable pattern, and morphological predictors. Results showed that the Random Forest algorithm outperformed Support Vector Machines in identifying readability levels of texts using a hybrid combination of traditional features (TRAD) and syllable pattern features (SYLL) with an accuracy of 0.661. Future directions of the study can focus on inclusion of non text-based features such as audio from recorded readings of texts and psycholinguistic features such as age-of-acquisition of words can be explored to achieve a multi-modal approach in building the readability assessment model.

## Acknowledgments

The authors would like to thank the anonymous reviewers for their valuable feedback as well as Adarna House Inc. for allowing us to use their dataset for this study.